# GAN-Based Single-Stage Defense for Traffic Sign Classification Under Adversarial Patch

Abyad Enan[1,*], and Mashrur Chowdhury[1]

**Abstract –** Computer vision plays a critical role in ensuring the safe navigation of autonomous vehicles (AVs). An AV perception module facilitates safe navigation. This module enables AVs to recognize traffic signs, traffic lights, and various road users. However, the perception module is vulnerable to adversarial attacks, which can compromise its accuracy and reliability. One such attack is the adversarial patch attack (APA), an attack in which an adversary strategically places a specially crafted sticker on an object to deceive object classifiers. Such an APA can cause AVs to misclassify traffic signs, leading to catastrophic incidents. To enhance the security of an AV perception system against APAs, this study develops a Generative Adversarial Network (GAN)-based single-stage defense strategy for traffic sign classification. This approach is tailored to defend against APAs across different classes of traffic signs, without prior knowledge of a patch's design, and is effective against patches of varying sizes. In addition, our single-stage defense is computationally efficient, requiring significantly lower computation time than existing multi-stage defenses, making it suitable for real-time deployment in autonomous driving systems. Compared to a classifier without any defense mechanism, our experimental analysis demonstrates that the defense strategy presented in this paper improves our classifier's accuracy under APA conditions by up to 90% considering traffic sign classes considered in this study. and overall classification accuracy enhances by 55% for all traffic signs considered in this study. Our defense strategy is model agnostic, making it applicable to any traffic sign classifier, regardless of the underlying classification model.

## 1. Introduction

With the advancement of artificial intelligence, computer vision has become a critical component of intelligent transportation systems (ITS). It plays a fundamental role in enhancing both safety and mobility, particularly in connected and autonomous vehicle (CAV) technologies. Computer vision is widely applied in various ITS domains, including traffic surveillance [1], smart toll collection [2], video-based basic safety message generation [3], smart parking management [4], and other object detection-based applications.

One of the most significant applications of computer vision in ITS is in autonomous vehicles (AVs) [5, 6]. AVs rely heavily on computer vision techniques for safe navigation by enabling the perception module to recognize traffic signs, road markings, traffic lights, and other road users. The perception modules are primarily developed using modern machine learning (ML) and deep learning (DL) techniques. However, adversarial machine learning has emerged as a threat, designed to exploit and deceive ML/DL models. In adversarial machine learning, adversaries analyze the behavior of ML/DL models to understand their decision-making process and identify vulnerabilities [7]. By manipulating the input data, adversaries can alter the model's predictions, leading to incorrect classifications [8]. Consequently, the perception module of AVs, which relies on ML/DL techniques, remains susceptible to cyberattacks [9].

A specific type of adversarial attack, known as the adversarial patch attack (APA), involves an adversary placing a specially crafted physical patch, generated using adversarial machine learning, on an object to mislead object classifiers [10]. These adversarial patches are typically small, occupying only a minimal portion of the object to maintain stealth and are visually inconspicuous to humans. Despite their small size, adversaries carefully designed these patches with specific patterns that can deceive ML/DL models in recognizing objects with high success rates [11]. When placed on a traffic sign, an adversarial patch can cause an AV's perception module to misclassify the sign, potentially leading to catastrophic consequences.

Fig. 1 interprets the concept of an APA targeting a traffic sign classifier. An adversary observes the inputs and outputs of a traffic sign classifier to craft an adversarial patch that induces misclassification. The adversary then places an adversarial patch on a traffic sign to manipulate a classifier's predictions. Fig. 1 illustrates an example in which the adversary applies the patch to a "pedestrian crossing" sign, causing the classifier to misidentify it as a "speed limit-45 sign". However, the patch can be applied to various traffic signs, depending on the adversary's objective and the specific signs considered during a patch generation process.

The misclassification of traffic signs can have severe consequences for AVs. For instance, if an AV misclassifies a "stop sign" as a "speed limit-45" sign, it could result in a life-threatening incident. Therefore, effective countermeasures must be implemented to safeguard AVs against such adversarial threats. Numerous works have been done to defend APA [12–

[*]Corresponding author: Abyad Enan (aenan@clemson.edu)

[1]Glenn Department of Civil Engineering, Clemson University, Clemson, SC 29634, USA

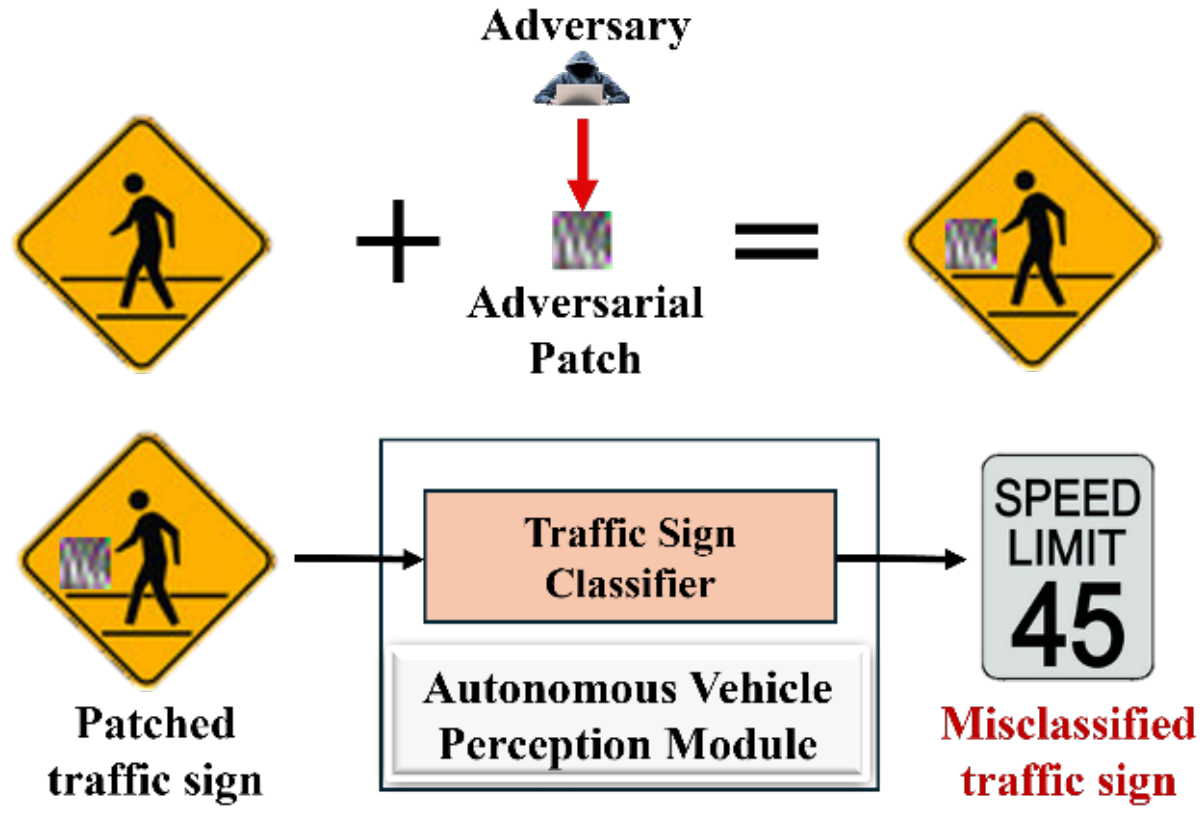

**Fig. 1** Adversarial patch attack on a traffic sign classifier of the perception module of an autonomous vehicle.

14]. However, existing defense methods involve multiple stages: adversarial patch detection followed by patch effect nullification [12, 14], making them computationally expensive and infeasible for real-time implementation for applications, such as autonomous driving. Another limitation is that existing methods mask the adversarial-patched areas [12, 14], and while doing so, they may hide other areas not covered by adversarial patches that contain important features, potentially hiding important object features, or they may fail to detect adversarial patches, reducing the accuracy of vision-based object classifiers.

To address these limitations, we develop a Generative Adversarial Network (GAN)-based single-stage defense strategy to protect traffic sign classifiers in AVs from APAs. The advantage of our GAN-based defense is that it enables a single-stage approach to restore patch-free images without requiring a separate patch-detection step, unlike methods in the literature, making it computationally efficient for real-time implementation in AVs. Moreover, a single generator can effectively reconstruct patch-free images across different traffic sign classes. This study addresses the research gaps by developing a robust defense mechanism for AVs for traffic sign classification. We evaluate our method against state-of-the-art methods and find that it outperforms them across detection accuracy, precision, recall, F1-score, and computation time. This defense strategy can be seamlessly integrated into a perception module of AVs, enhancing their resilience against APAs. The key contributions of this study are as follows:

- A single-stage defense strategy against APAs for traffic sign classification in AVs using GANs.
- Formulation of a loss function, incorporating a weighted combination of multiple losses, to train the GAN for reconstructing patch-free images while preserving essential classifiable features.
- A GAN-based defense strategy capable of reconstructing adversarial patch-free images across multiple classes of traffic signs using a single generator.
- An APA resilient traffic sign classification system that performs well for diverse sizes of adversarial patches without having any prior knowledge about the patch design.
- A GAN-based defense strategy that is robust and reproducible for real-world implementations.

The rest of the manuscript is structured as follows: Section 2 reviews related studies. Section 3 discusses the attack model, Section 4 explains the defense strategy, Section 5 narrates the research method, Section 8 covers the analysis and results, and Section 7 evaluates the defense strategy on a benchmark dataset. Finally, Section 8 draws conclusions from our study.

## 2. **Related Work**

Numerous works have been done to develop adversarial patch attacks in computer vision; however, very few studies have focused on developing defense strategies against APA.

One of the earliest approaches to mitigating adversarial patch attacks is through saliency map analysis, which helps detect abnormally high activation areas within an image. Hayes et al. (2018) developed a method that leverages watermark removal techniques by identifying suspicious regions and masking them [15]. Naseer et al. (2019) developed the Local Gradient Smoothing (LGS) technique, which suppresses irregular gradients caused by adversarial patches. However, these methods come with several disadvantages. This method's false positive rate is very high, as saliency maps may misidentify normal high-contrast regions (such as shadows, textures, or naturally salient objects) as adversarial patches. This can lead to unnecessary modifications to non-adversarial images, reducing the model's effectiveness. Besides, computing saliency maps in real-time can be computationally expensive, making these defenses less suitable for high-speed, real-world applications such as autonomous driving or real-time security surveillance.

Another defense strategy involves restricting the model's receptive field size to limit the influence of an adversarial patch. Xiang et al. (2020) developed a PatchGuard (PG) that ensures that only small portions of an image contribute to the final classification, thus preventing localized adversarial patches from dominating predictions [16]. Later on, Xiang et al. (2021) developed PatchGuard++, an improvement over PG [17]. However, restricting the receptive field may make them struggle to understand larger, more complex patterns in images. This can degrade accuracy on clean, unperturbed data. Processing smaller receptive fields individually requires additional post-processing techniques (e.g., feature aggregation), increasing computational overhead. Besides, they are designed for a specific CNN architecture and may lack transferability.

In [18], the author developed a text-guided diffusion model to detect and counter adversarial patch attacks by leveraging the Adversarial Anomaly Perception (AAP) phenomenon, enabling accurate patch localization and restoration within a unified framework. However, the approach relies on diffusion models, which are computationally expensive, particularly for real-time applications.

In [12], the authors developed a Patch-Agnostic Defense (PAD) method that detects and removes adversarial patch attacks by leveraging semantic independence and spatial heterogeneity, enabling effective patch localization without prior attack knowledge or additional training. However, the method is computationally expensive as it involves multiple steps, including mutual information computation, recompression-based analysis, and segmentation model integration, which may introduce latency in real-time applications.

Mao et al. (2024), developed a universal defense filter (UDFilter) against adversarial patch attacks that enhances robustness in person detection by overlaying a self-adaptive defense filter on input images, mitigating the impact of adversarial patch attacks without requiring modifications to pre-trained object detectors [19]. Although UDFilter avoids modifying the detection model itself, generating an effective universal filter involves an iterative adversarial training process, requiring additional computational resources.

Chen et al. (2023), in [13], developed a two-stage defense strategy, Jujutsu, against APAs that first detect adversarial patches and then mask the patched area. Finally, using GANs, the masked areas are reconstructed. However, Jujutsu requires significant computational overhead due to the multi-step detection and mitigation processes. The overhead is primarily caused by saliency maps computation, taking around 340 milliseconds (ms), which makes Jujutsu less efficient for real-time applications [13].

Zhang et al. (2022) proposed a defense strategy for traffic signs against APAs, where they considered 50 different mask patterns with black and white rectangular tapes as adversarial patches [20]. To defend against APAs, the authors trained a traffic sign classifier on traffic sign images with 50 predefined adversarial patches applied to each image. However, this method lacks generalization, as it is applicable only to APAs with known adversarial patches and may not adapt to unseen attacks [20].

## 3. **Attack Model**

This study focuses on APAs, where a small, learnable patch can be placed anywhere on a traffic sign to mislead a traffic sign classifier. Unlike conventional adversarial perturbations that modify each pixel of an entire image [8], APAs target only a small region of the object. These patches can be applied either digitally or physically, for example, by printing the adversarial patch and placing it on the top of the target object, posing a potential security threat. The attack model considered in this study is adapted from a benchmark patch generation method presented in a prior work on gradient-based attacks by Brown et al. (2018 ) in [10]. We consider this attack model for performing digital attacks in this study. However, it can also be used for physical attacks by printing adversarial patches and placing them on the tops of traffic signs [10].

The APA performed in this study follows a white-box attack model, where the adversary has complete knowledge of the DL model, including its parameters, gradients, and class probabilities. The objective is to generate an adversarial patch that, when placed on any targeted traffic sign, causes the classifier to misidentify it. The key components of the threat model are as follows: (a) Knowledge about the DL Model: The adversary has complete knowledge of the neural network architecture, model parameters, and gradient information of the traffic sign classifier; (b) Access to Training Data: The adversary can utilize the training dataset, enabling optimization of the adversarial patch across multiple images; (c) No Modification of Model Weights: While the adversary has knowledge of the classifier's DL model, they cannot modify its parameters, ensuring that the attack relies solely on adversarial patch manipulation.

The patch generation problem is formulated as an optimization problem in which a small, randomly initialized patch is placed at multiple positions on clean traffic sign images. The patch is then optimized so that, when applied to the images regardless of patch position, the DL model misclassifies the traffic sign images with high confidence. The optimization process involves computing the cross-entropy loss and iteratively updating the patch using the Adam optimizer. The key steps for generating the adversarial patch are as follows:

- Patch Initialization: A patch of a predefined size is randomly initialized.
- Random Placement: An initialized patch is applied at random positions in each training image.
- Forward Pass Through the Classifier: The patched images are fed into the traffic sign classifier to obtain predictions.
- Loss Calculation: Cross-entropy is computed to maximize the probability for misclassification when the adversarial patch is present.
- Patch Update: Gradients are computed, and the patch is updated using the Adam optimizer.
- Validation: The subject patch is evaluated on a validation dataset to assess its effectiveness in every epoch.

This adversarial patch generation process ensures that the patch induces effective misclassification in order to highlight the vulnerability of traffic sign classifiers to APAs. Following the steps, adversarial patches of varying sizes are generated to evaluate the effectiveness of our defense strategy across different patch sizes.

Although the attack in this study is a white-box attack, in which the patch is generated with prior knowledge of the DL model used for traffic sign classification, the adversarial patch can also successfully mislead other DL-based traffic sign classifiers for which the adversary has no prior knowledge of their architectures or parameters. That means the APAs we performed in this study exhibit transferability. The transferability of the patch is demonstrated in Section 6, where experimental results confirm that an adversarial patch, initially crafted for a specific DL model, can deceive other models as well. This transferability suggests that adversarial patches exploit universal weaknesses in convolutional neural networks (CNNs). By leveraging gradient-based optimization, the patch manipulates CNN feature representations in a model-agnostic manner, enabling it to deceive even models about which the adversary has no knowledge, such as those for which the adversary has no direct access to the traffic sign classifier's model details.

4. **Defense Strategy**

We develop a GAN-based single-stage defense strategy to protect the traffic sign classifier of the AVs perception module. The GAN is trained so that when a patched traffic sign image is fed to the generator, it produces a patch-free version of the image with meaningful features that are classifiable by the traffic sign classifier. The strategy suppresses the threat that an APA might pose in a single step. The idea is illustrated in Fig. 2.

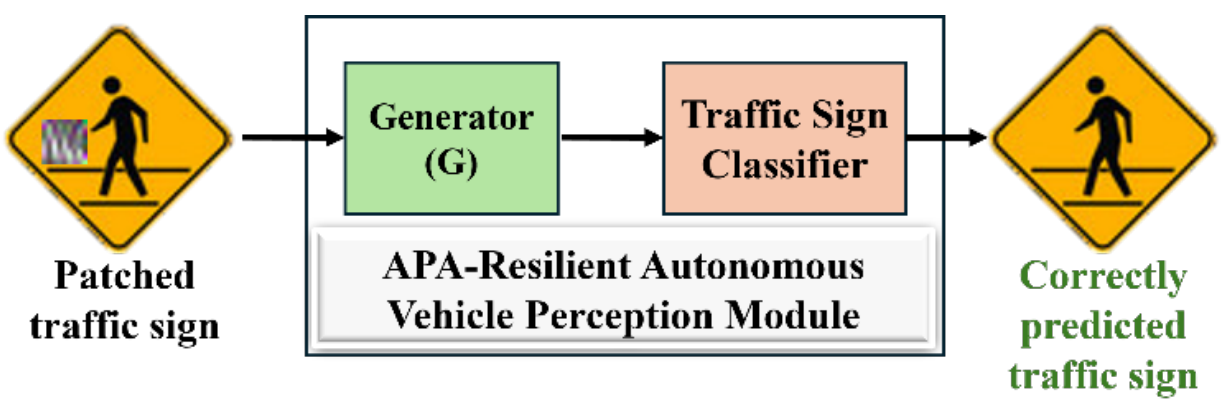

**Fig. 2** GAN-based single-stage defense strategy where the generator generates adversarial patch-free traffic sign images.

4.1 **GAN Architecture**

The GAN consists of two neural networks: a generator and a discriminator. The generator restores patched traffic sign images, while the discriminator distinguishes real from generated (fake) images. The Generator utilizes an encoder-decoder structure with attention mechanisms to improve feature learning and reconstruction quality. It is designed to remove physical patches (i.e., occlusions) from traffic sign images and reconstruct the original (i.e., patch-free) appearance.

The encoder extracts hierarchical feature representations while progressively downsampling the spatial resolution. Convolutional layers progressively reduce the spatial size while increasing the number of channels, helping the model extract meaningful features. Attention blocks allow the network to focus on important regions of the input, helping to reconstruct occluded patched areas by learning long-range dependencies. The decoder upsamples the feature maps to reconstruct the original input. Deconvolution layers, transpose convolutions, upsample the feature maps back to the original resolution (32 × 32), where attention blocks help refine details by maintaining spatial relationships and correcting patched areas.

The Generator architecture is given in Table 1. For the convolutional layers Conv1, Conv2, Deconv1, and Deconv2, stride = 2, padding = 1, for Conv1, Conv2, and Deconv1 ReLU activation function, and for Deconv2 Sigmoid activation function is used. The final Sigmoid activation normalizes pixel values to the range [0, 1].

The discriminator is a binary classifier (as shown in Table 2) that determines whether an image is real or generated. For both convolutional layers, stride = 2 and padding =1, and the LeakyReLU(0.2) activation function is used. The flatten layer converts feature maps into a fully connected representation, and the final fully connected layer uses the Sigmoid activation function to output 0 for fake or 1 for real.

Unlike traditional GANs that generate images from random noise, our generator takes a patched image as input and learns to restore the missing information through a conditional GAN.

**Table 1** Components of the generator architecture

| **Encoder** | | | | |
|---|---|---|---|---|
| **Layer Name** | **Type** | **Input Dimension** | **Output Dimension** | **Kernel** |
| Conv1 | Conv2D | (3, 32, 32) | (64, 16, 16) | 4×4 |
| Attention1 (conv1) | Conv2D (1×1) | (64, 16, 16) | (8, 16, 16) | 1×1 |
| Attention1 (conv2) | Conv2D (1×1) | (64, 16, 16) | (8, 16, 16) | 1×1 |
| Attention1 (conv3) | Conv2D (1×1) | (64, 16, 16) | (64, 16, 16) | 1×1 |
| Conv2 | Conv2D | (64, 16, 16) | (128, 8, 8) | 4×4 |
| Attention2 (conv1) | Conv2D (1×1) | (128, 8, 8) | (16, 8, 8) | 1×1 |
| Attention2 (conv2) | Conv2D (1×1) | (128, 8, 8) | (16, 8, 8) | 1×1 |
| Attention2 (conv3) | Conv2D (1×1) | (128, 8, 8) | (128, 8, 8) | 1×1 |
| **Decoder** | | | | |
| **Layer** | **Type** | **Input Dimension** | **Output Dimension** | **Kernel** |
| Deconv1 | ConvTranspose2D | (128, 8, 8) | (64, 16, 16) | 4×4 |
| Attention3 (conv1) | Conv2D (1×1) | (64, 16, 16) | (8, 16, 16) | 1×1 |
| Attention3 (conv2) | Conv2D (1×1) | (64, 16, 16) | (8, 16, 16) | 1×1 |
| Attention3 (conv3) | Conv2D (1×1) | (64, 16, 16) | (64, 16, 16) | 1×1 |
| Deconv2 | ConvTranspose2D | (64, 16, 16) | (3, 32, 32) | 4×4 |

**Table 2** Components of the discriminator architecture

| **Layer Name** | **Type** | **Input Dimension** | **Output Dimension** | **Kernel** |
|---|---|---|---|---|
| Conv1 | Conv2D | (3, 32, 32) | (64, 16, 16) | 4×4 |
| Conv2 | Conv2D | (64, 16, 16) | (128, 8, 8) | 4×4 |
| Flatten | Flatten | (128, 8, 8) | (8192) | - |
| FC | Fully Connected | (8192) | (1) | - |

4.2 **GAN Train for APA Defense**

The GAN training stage involves training both the generator (G) and discriminator (D) together, especially the generator, to generate patch-free traffic sign images with meaningful features for the classifier to make the images classifiable. Basically, we train the generator to extract important features of the traffic signs and generate images with those features. To achieve this, we use two datasets of images. The first dataset is the dataset with clean, attack-free traffic sign images, with which the traffic sign classifier is already trained. The other dataset is created by placing random patches, with random sizes, in random positions of the clean traffic sign images. These patches are completely random, not adversarial crafted patches, and not related to the adversarial patches generated for this study. They are just small (compared to the traffic sign in size), colorful image patches with random pixel values. The purpose of using random-sized, randomly generated, randomly placed patches on the traffic sign images is that, in practice, an adversary uses an adversarial patch about which the perception module of AVs has zero knowledge, and we want the generator to be trained in a way that it is capable of removing any type of unwanted patches placed at the top of the traffic signs regardless the texture, structure, and position.

Then, based on the generator's generated images and the respective clean images, the true label of the input patched image, and the classifier's predicted label towards the generator's generated image, we calculate different losses to improve the performance of the generator. The training process is illustrated in Fig. 3, and losses are represented in the following equations:

$$\mathcal{L}_D = \mathcal{L}_{D,adv} = -\frac{1}{2}\left[E_{x\sim p_{data}(x)} \log D(x) + E_{\tilde{x}\sim p_{data}(\tilde{x})} \log(1 - D(G(\tilde{x})))\right] \quad (1)$$

$$\mathcal{L}_{G,adv} = -E_{\tilde{x}\sim p_{data}(\tilde{x})} \log(D(G(\tilde{x}))) \quad (2)$$

$$\mathcal{L}_{recon} = E[\|G(\tilde{x}) - x\|^2] \quad (3)$$

$$\mathcal{L}_{percep} = \sum_l \|\varphi_l(G(\tilde{x})) - \varphi_l(x)\|^2 \quad (4)$$

$$\mathcal{L}_{class} = -\sum y_i \log C(G(\tilde{x})) \quad (5)$$

$$\mathcal{L}_G = \mathcal{L}_{G,adv} + 10 \times \mathcal{L}_{recon} + 0.1 \times \mathcal{L}_{percep} + 5 \times \mathcal{L}_{class} \quad (6)$$

Here, $x\sim p_{data}(x)$ represents real images from the dataset, $\tilde{x}\sim p_{data}(\tilde{x})$ represents patched images from the patched dataset, $G(\tilde{x})$ is the reconstructed image, $D(x)$ is the discriminator's probability that $x$ is real, $D(G(\tilde{x}))$ is the probability that the discriminator thinks the reconstructed image is real, $\varphi_l(.)$ extracts features from the layer $l$ of the classifier, and $C(G(\tilde{x}))$ is the classification outcome from the traffic sign classifier for the reconstructed image $G(\tilde{x})$.

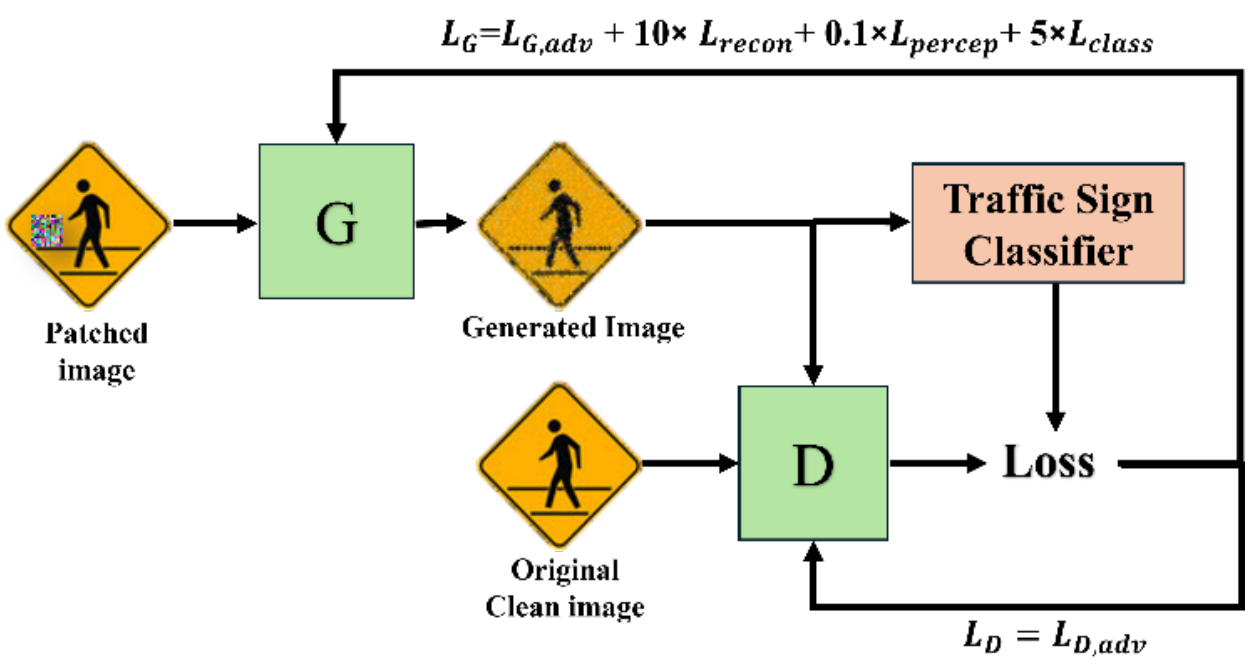

**Fig. 3** GAN training process and loss calculation.

Equation (1) computes the total loss of the discriminator equivalent to the discriminator adversarial loss, which is based on the Binary Cross-Entropy (BCE) loss. The discriminator is trained to distinguish real images from fake images reconstructed by the generator. Equation (2) is the adversarial loss of the generator, which is minimized when the generator fools the discriminator. $\mathcal{L}_{D,adv}$ pushes the generator to produce images to fool the discriminator into classifying them as real. Equation (3) represents the reconstruction loss, which is the pixel-wise difference between the original patch-free and the reconstructed images, ensuring that the reconstructed image closely matches the patch-free clean image, even if the input image is patched. Equation (4) represents the perceptual loss that ensures that feature maps of the reconstructed images resemble the original patch-free images using a feature extractor from the classifier. Equation (5) is the classification loss, uses Cross-Entropy loss, and ensures that the reconstructed image preserves class-specific features to correctly classify it by the classifier. Equation (6) is the total generator loss, which is the weighted combination of the losses in (2)-(5). This weighted combination of the loss function enables the generator to balance realism, structure, and class consistency in the reconstructed patch-free images during training.

## 5. **Research Method**

This section discusses the research method of this study, including the preparation of the traffic sign image dataset, the training and testing of the traffic sign classifier, and the selection of the GAN model.

### 5.1 **Traffic sign image dataset preparation**

In this study, we develop a custom traffic sign dataset, which is a subset of the LISA traffic sign dataset. The LISA dataset consists of real-world traffic sign images representing 49 U.S. traffic sign types. The images in this dataset were captured using dash cameras mounted on vehicles driving through San Diego, California. The traffic sign annotations vary in size, ranging from 6 × 6 to 167 × 168 pixels, and include both color and grayscale images. The dataset comprises images captured under diverse weather conditions, lighting conditions, and viewing angles. Due to these variations, the LISA dataset is widely used in training autonomous driving systems and other traffic sign recognition applications.

For this study, we select a subset of five different traffic sign classes from the LISA dataset, covering Regulatory and Warning categories: Pedestrian Crossing (Ped-Xing), School Zone, Signal Ahead, Speed Limit-45, and Stop Sign. For each traffic sign, we use images of a uniform size of 32 × 32 pixels with three color channels (red, green, and blue). The customized dataset contains 625 images in total, with 125 per class. The dataset is split into training, validation, and testing subsets, allocating 75, 25, and 25 images per class, respectively. Sample traffic sign images from each class are presented in Fig. 4.

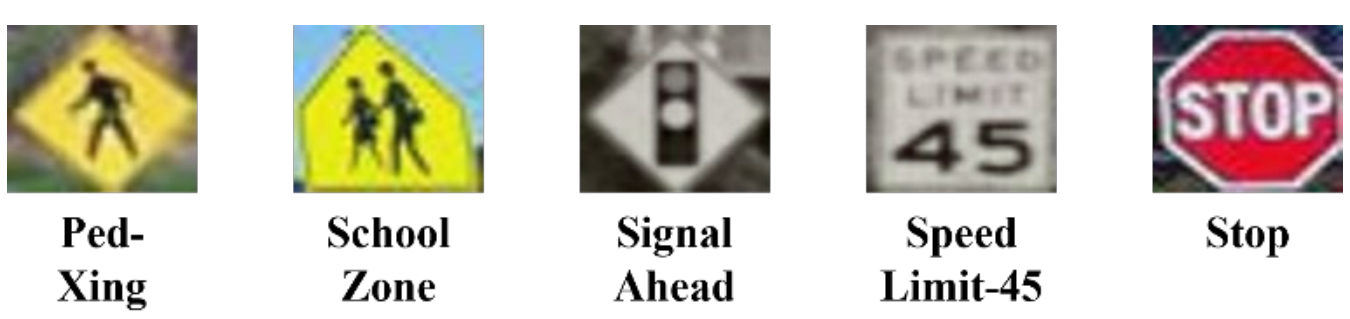

**Fig. 4** Five classes of traffic signs considered in this study.

### 5.2 **Traffic sign classifier training and testing**

For the traffic sign classifier, we select the ResNet-50 model, a 50-layer deep convolutional neural network (CNN) from the Residual Network (ResNet) family for the traffic sign classifier. The ResNet-50 is widely used for object detection, image classification, and various computer vision tasks due to its robust feature extraction capabilities. In this study, ResNet-50 is trained on the custom traffic sign dataset, and its trained weights and model parameters are utilized to generate adversarial patches of varying sizes, which are later used to evaluate the effectiveness of our defense strategy.

Beyond the primary ResNet-50-based traffic sign classifier, we employ an additional classifier to assess the transferability of adversarial patches. For this purpose, we use Inception-V3, another deep convolutional neural network. The Inception-V3 is also trained on the custom traffic sign dataset, but its role is solely to evaluate whether the adversarial patch attack (APA) performed in this study remains effective against alternative traffic sign classifiers utilizing different deep learning models, especially when an adversary has zero knowledge of their architecture or parameters. To assess the transferability and effectiveness of the attack, we measure the Inception-V3 model's classification accuracy before and after the attack.

### 5.3 **Adversarial patch generation**

We generate adversarial patches of varying sizes using the method described in Section 3. The process begins by initializing a patch of predefined size, which is then placed at different positions on the traffic signs. The patch is optimized to mislead the classifier into predicting incorrect traffic sign classes. Following this approach, we generate three different adversarial patches with sizes of 3 × 3, 4 × 4, and 5 × 5 pixels.

The primary motivation for experimenting with different patch sizes is to evaluate the performance of the defense strategy under varying attack conditions. While no strict size constraints for adversarial patches have been established in the literature, adversarial patches are generally expected to be small relative to the target object to maintain stealthiness. Smaller patches are harder to detect, whereas larger patches tend to be more effective at misleading classifiers. However, if an adversarial patch covers too much of the image, the attack may become impractical or trivial, as it could simply occlude the entire object rather than exploit model vulnerabilities.

### 5.4 **Best generator selection for defense**

The GAN is trained following the strategy outlined in Section 4. During each epoch, losses are computed, and the performance of both the generator and the discriminator is iteratively improved. The generator is provided with training images that have been randomly patched with randomly generated patches at random positions. Its objective is to reconstruct images by removing patches while preserving the classifiable and meaningful features of patch-free traffic signs. At the end of each training epoch, validation images from the dataset are patched with random patches, as described in Section 4, and are subsequently fed into the generator for reconstruction. The reconstructed images are then passed through the traffic sign classifier, and the validation accuracy is determined by comparing the classifier's predictions with the ground truth labels.

Throughout the training process, two key metrics are computed and observed at each epoch: Structural Similarity Index Measure (SSIM) of the reconstructed images, which evaluates how well the generator restores the original clean traffic sign features, and validation accuracy of the classifier on the reconstructed images, which measures the classifier's ability to correctly classify patch-free images generated by the GAN. After training, the generator with SSIM > 0.5 and the highest validation accuracy on the classifier is selected as the optimal generator for the defense strategy.

### 5.5 **Performance comparison with existing APA defense methods**

We compare the performance of our GAN-based defense strategy with existing baselines, including an adversarial training method, and the patch-agnostic defense (PAD) [12] method developed by Jing et al. Adversarial training enhances the robustness of DL models by exposing them to adversarial samples during training. Specifically, for this baseline, we train the ResNet-50 traffic sign classifier using training images that are randomly patched with randomly generated patches placed at random positions, i.e., the same patched images used to train the GAN in Section 5.4.

PAD [12], developed by Jing et al., is an APA defense strategy that does not require prior knowledge of the patch's size, shape, location, or appearance. It employs a three-stage, training-free pipeline consisting of Dual-Characteristic Localization, refinement using the Segment Anything Model (SAM), and mask-based patch removal to restore classifier accuracy [12]. For this method, we use the authors' publicly available implementation and evaluate the defended images with the ResNet-50 classifier. For performance assessment of all defense methods, we report precision, class-wise accuracy (recall), F1-score, overall classification accuracy, and computation time.

## 6. **Analysis and Results**

This section discusses the analysis performed and the results before and after incorporating the defense strategy to evaluate the strategy developed in this study. We also compare the performance of our approach with the existing benchmarks.

Each method is executed five times, and the average performance metrics are reported. Since 125 images are used per class, where 25 are designated for testing in each run, to ensure that every image serves as a test sample at least once across the five runs, we apply a circular split to the dataset, which motivates running the models five times.

### 6.1 **Baseline performance of the traffic sign classifier**

The traffic sign classifier, implemented using the ResNet-50 model, is trained and evaluated on the customized real-world traffic sign dataset prepared for this study. Fig. 5 presents the baseline accuracy for each traffic sign class, which is the recall for each class. When the classifier operates without an APA, the overall baseline accuracy is 97.76% across all five traffic sign classes.

### 6.2 **Performance of the traffic sign classifier under APAs**

We generate adversarial patches of different sizes and apply them at random positions on the test images to evaluate the classifier's performance under attack conditions. For each patch size, we analyze the decline in classification accuracy across traffic sign classes. The results are summarized in Fig. 6.

Fig. 6 demonstrates a significant drop in accuracy across all traffic sign classes after applying APAs. The results in Table 3 indicate that the average accuracy for the pedestrian crossing (ped-xing) sign drops to nearly 0%, while the stop sign's accuracy falls below 20%. Other traffic signs also experience a substantial decline in average accuracy. Overall, across all five traffic sign classes, the classifier's accuracy under attack conditions drops to 39.25%, as shown in Table 4, highlighting the effectiveness of the APA performed in this study.

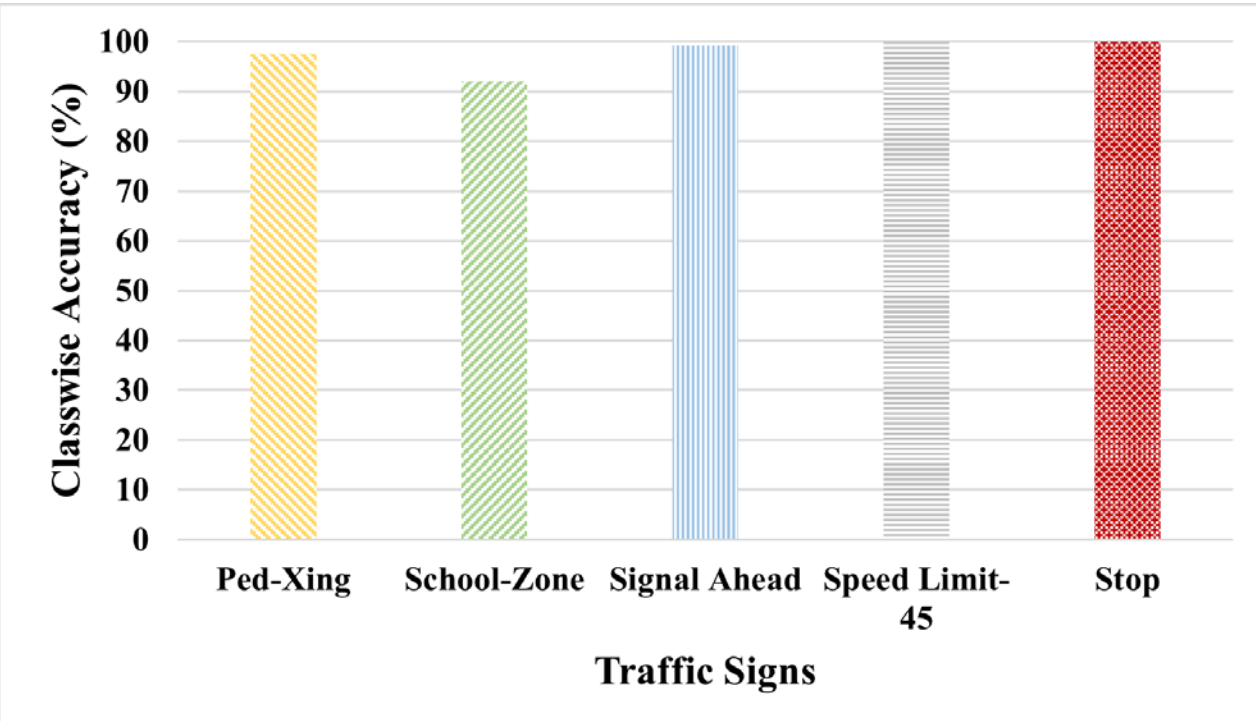

**Fig. 5** Baseline accuracy of the traffic sign classifier, ResNet-50 model for individual traffic sign classes.

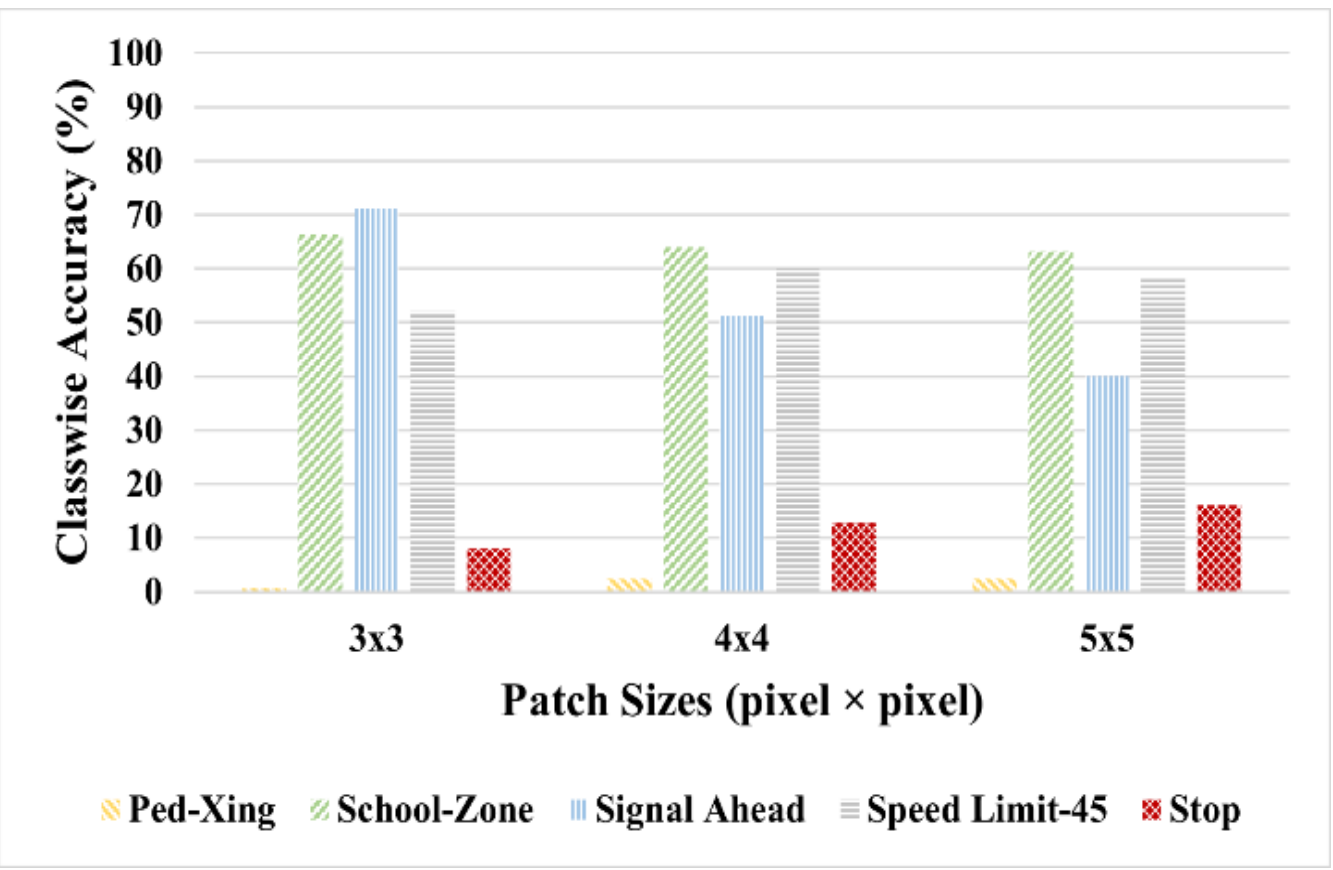

**Fig. 6** Accuracy of the traffic sign classifier, ResNet-50 model for individual traffic sign classes, under attack conditions for different patch sizes of APAs.

6.3 **APA transferability check**

To evaluate the transferability of the APAs generated in this study, we test another traffic sign classifier trained with a different DL model, Inception-V3, for which the adversary has no prior knowledge. The goal of this transferability analysis is to determine whether the APAs generated in this study remain effective against different DL models, even when the attacker lacks any information about the target model's architecture or parameters.

Fig. 7 presents the results of the transferability test. Fig. 7(a) presents the accuracy of Inception-V3 under baseline conditions, whereas Fig. 7(b) highlights that the accuracy falls under attack conditions. The findings indicate that the APAs in this study successfully deceive the Inception-V3 model, demonstrating their ability to mislead black-box models without requiring prior knowledge of their internal workings. This confirms that the attack is model-agnostic and generalizable, posing a significant threat to a wide range of traffic sign classifiers.

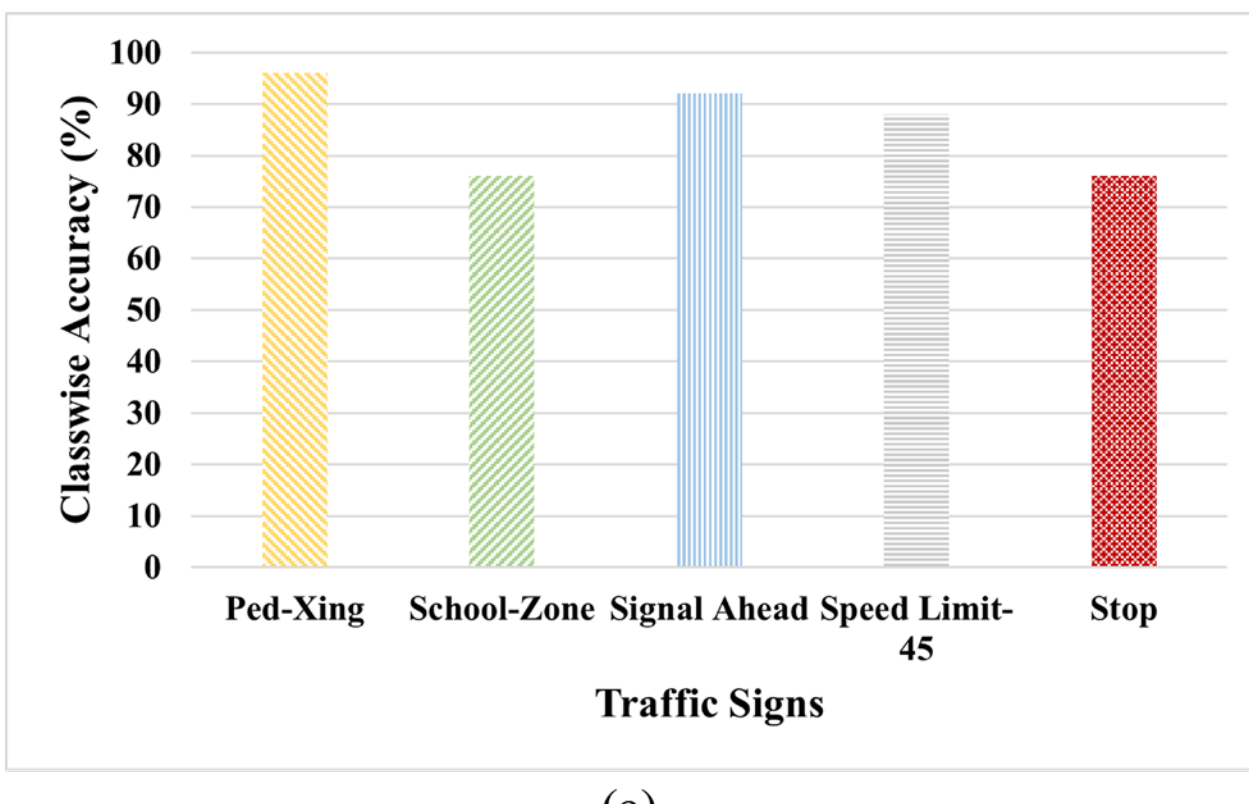

(a)

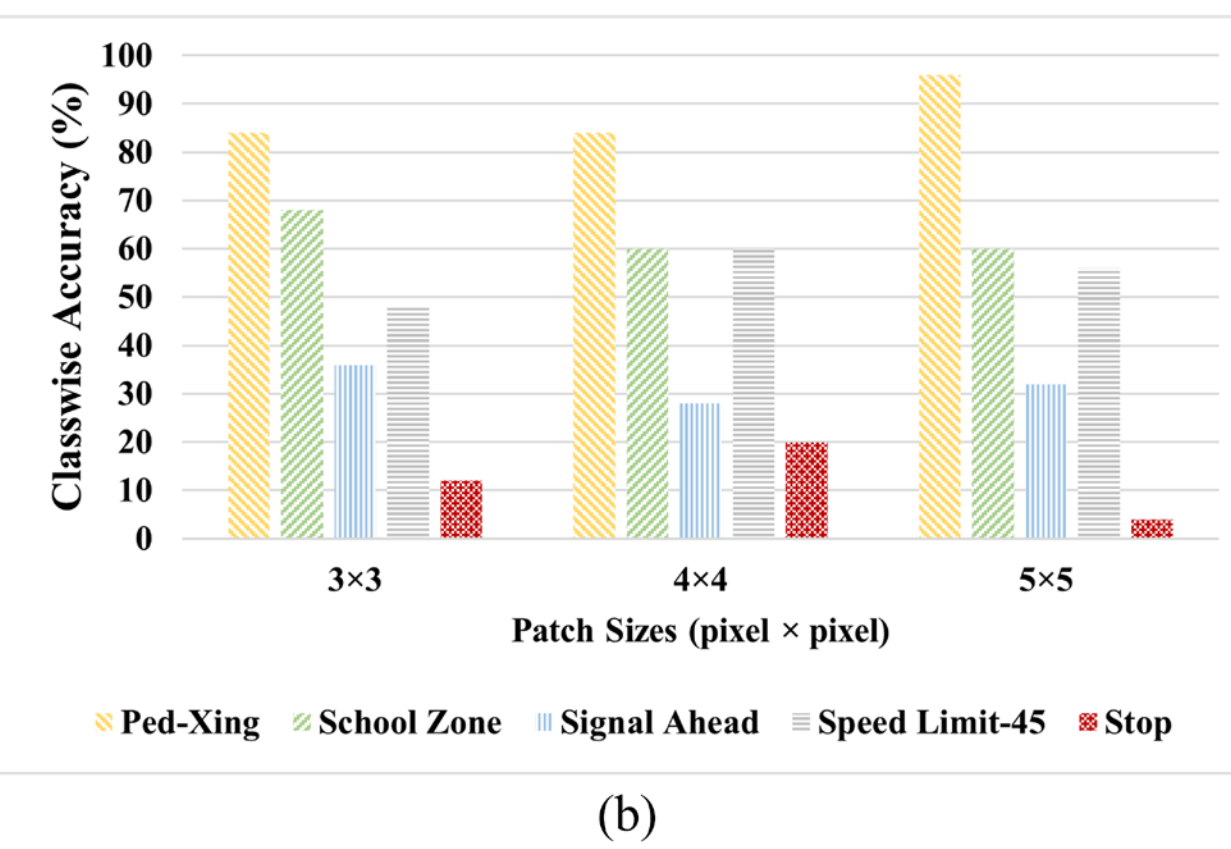

(b)

**Fig. 7** Accuracy of an Inception-V3 model-based traffic sign classifier (a) without attack condition and (b) under APAs performed using the similar patches generated to target the ResNet-50 model

6.4 **Evaluation of GAN-based single-stage defense strategy**

To evaluate the effectiveness of our defense strategy, we apply adversarial patches, generated using the method described in Section 3, to test traffic sign images at random positions. The adversarial-patched images are fed to the generator for reconstruction. The authors note that the adversarial patches used in the test images are completely unknown to the generator. The reconstructed images are subsequently fed into the ResNet-50 traffic sign classifier for classification. The results are summarized in Fig. 8.

Fig. 8 reveals a massive improvement in classification accuracy after applying the defense strategy. Specifically, the average accuracy for the ped-xing increases from 1.87% to 92%, for stop sign from 12.27% to 91.11%, for signal ahead from 54.13% to 94.22%, for speed-limit-45 from 56.8% to 95.56%, and for school zone from 64.53% to 94.22%.

Table 3 presents the class-wise accuracy (recall) of the classifier, under attack and after GAN-based defense is incorporated for different sizes of adversarial patches. We see that the average accuracy (recall) increases for the ped-xing is 89.69% which is the highest increase after the defense among all other signs. For the other traffic signs, the average accuracy increases by 78.84% for the stop sign, 40.69% for the signal ahead, 38.76% for the speed-limit-45, and 29.69% for the school zone.

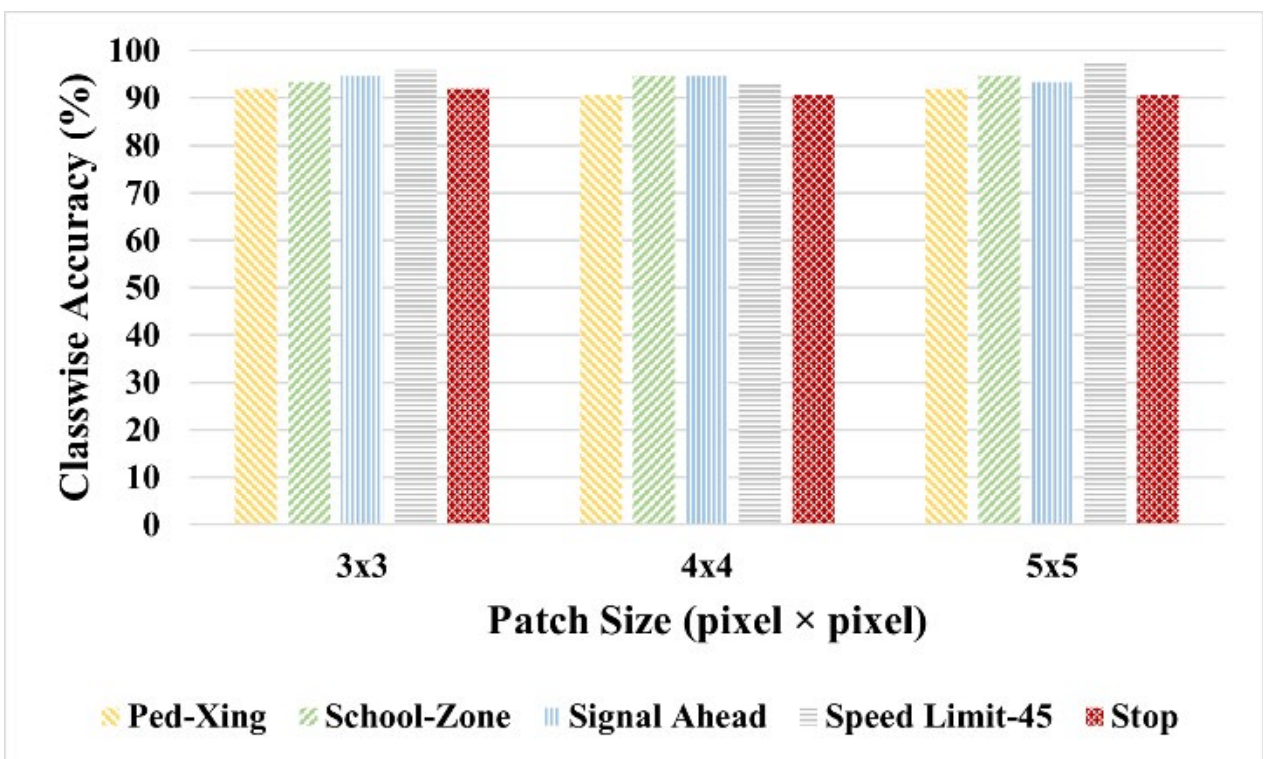

**Fig. 8** Accuracy of the traffic sign classifier, ResNet-50 model for individual traffic sign classes, under attack for different patch sizes of APAs after the defense strategy is incorporated.

Overall, across all five traffic sign classes, the traffic sign classifier's accuracy after employing the defense strategy reaches 93.33%, representing a 55.41% accuracy improvement over a classifier without any defense mechanism, as shown in Table 4.

**Table 3** Class-wise accuracy (recall) for different classes of traffic signs

| Traffic Signs | Patch Size (px × px) | Acc UA (%) | Acc AD (%) | Acc Increased (%) | Avg Acc Increased (%) |
|---|---|---|---|---|---|
| Ped-Xing | 3×3 | 0.8 | 92 | 91.2 | 89.69 |
| | 4×4 | 2.4 | 90.67 | 88.27 | |
| | 5×5 | 2.4 | 92 | 89.6 | |
| School Zone | 3×3 | 66.4 | 93.33 | 26.93 | 29.69 |
| | 4×4 | 64 | 94.67 | 30.67 | |
| | 5×5 | 63.2 | 94.67 | 31.47 | |
| Signal Ahead | 3×3 | 71.2 | 94.67 | 23.47 | 40.09 |
| | 4×4 | 51.2 | 94.67 | 43.47 | |
| | 5×5 | 40 | 93.33 | 53.33 | |
| Speed Limit-45 | 3×3 | 52 | 96 | 44 | 38.76 |
| | 4×4 | 60 | 93.33 | 33.33 | |
| | 5×5 | 58.4 | 97.33 | 38.93 | |
| Stop | 3×3 | 8 | 92 | 84 | 78.84 |
| | 4×4 | 12.8 | 90.67 | 77.87 | |
| | 5×5 | 36 | 90.67 | 54.67 | |

Acc: Accuracy; UA: Under Attack; AD: After Defense; px: Pixel

**Table 4** Accuracy for different sizes of adversarial patches on traffic signs

| Patch Size (px × px) | Acc UA (%) | Avg Acc UA (%) | Acc AD (%) | Avg Acc AD (%) | Acc Increased AD (%) |
|---|---|---|---|---|---|
| 3×3 | 39.68 | 37.92 | 93.6 | 93.33 | 53.92 |
| 4×4 | 38.08 | | 92.8 | | 54.72 |
| 5×5 | 36 | | 93.6 | | 57.6 |

Acc: Accuracy; UA: Under Attack; AD: After Defense; px: Pixel

The results demonstrate the effectiveness of our defense strategy in mitigating APA across various patch sizes. Notably, the defense strategy successfully restores classification accuracy when confronted with adversarial patches that the GAN did not encounter during training. This underscores its robustness and generalizability in securing traffic sign classifiers against physical APAs.

The performance comparison with the existing baselines is presented in Table 5. Across all patch sizes (3×3, 4×4, and 5×5) and all traffic sign classes, our method consistently outperforms both Adversarial Training and PAD in terms of precision, recall, F1-score, and accuracy, while also showing substantially lower standard deviation, indicating better robustness and stability. The baseline methods exhibit low performance across all metrics, with notably large standard deviations that indicate unstable behavior across runs. In contrast, our method maintains high and stable accuracy, typically exceeding 90% across all classes and patch sizes, while achieving consistently strong precision, recall, and F1 scores with minimal variability.

Class-wise analysis further highlights that challenging categories such as Ped-Xing, School Zone, and Signal Ahead suffer substantial performance losses under adversarial training and PAD, whereas our approach preserves high precision, recall and F1 score, reflecting effective recovery of correct class predictions. Moreover, the lower standard deviations observed across repeated runs with our method underscore its robustness and reliability, suggesting reduced sensitivity to variations in patch placement and test splits.

The lower accuracy observed for PAD can be attributed to its limited ability to reliably localize the patched region and, consequently, to correctly mask only the adversarial area. As illustrated in Fig. 9, this often leads to the unintended suppression of semantically important object features, thereby reducing classification performance. In contrast, our method successfully restores patch-free image content and effectively inpaints the adversarial patched regions with meaningful pixel-level features, preserving the semantic integrity of the traffic sign.

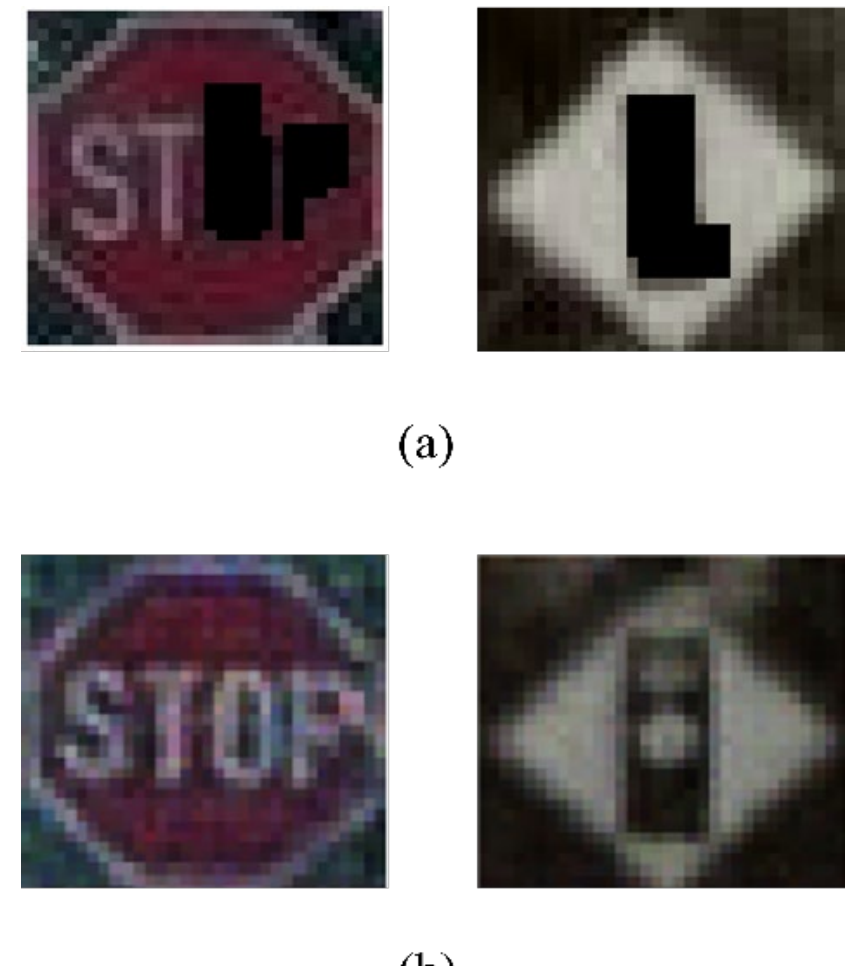

**Fig. 9** Sample APA defended images by (a) PAD method, and (b) our GAN-based defense method.

Overall, the results indicate that our defense provides a more resilient and stable solution for traffic sign classification under adversarial patch attacks, outperforming existing baselines.

**Table 5** Performance comparison against existing benchmarks

| Patch Size | Traffic Sign | Adversarial Training | | | | PAD [12] | | | | Ours | | | |
|---|---|---|---|---|---|---|---|---|---|---|---|---|---|
| | | **Prec (%) ± Std** | **Rec (%) ± Std** | **F1 (%) ± Std** | **Acc (%) ± Std** | **Prec (%) ± Std** | **Rec (%) ± Std** | **F1 (%) ± Std** | **Acc (%) ± Std** | **Prec (%) ± Std** | **Rec (%) ± Std** | **F1 (%) ± Std** | **Acc (%) ± Std** |
| 3×3 | Ped-Xing | 21.63 ± 17.75 | 12.8 ± 10.73 | 14.11 ± 10.86 | 36 ± 30.03 | 0.80 ± 1.79 | 10.00 ± 22.36 | 1.48 ± 3.31 | 41.98 ± 24.7 | 96.05 ± 3.85 | 92.00 ± 3.85 | 93.87 ± 0.25 | 93.6 ± 1.74 |
| | School Zone | 37.60 ± 16.78 | 79.2 ± 22.52 | 46.27 ± 10.39 | | 54.48 ± 20.03 | 61.89 ± 9.82 | 55.26 ± 10.00 | | 88.87 ± 6.94 | 93.33 ± 2.31 | 90.99 ± 4.66 | |
| | Signal Ahead | 55.66 ± 5.4 | 54.4 ± 25.71 | 51.69 ± 16.22 | | 45.18 ± 17.83 | 65.85 ± 21.79 | 49.36 ± 6.65 | | 93.57 ± 4.06 | 94.67 ± 2.31 | 94.06 ± 1.85 | |
| | Speed Limit-45 | 31.47 ± 42.99 | 8.8 ± 15.34 | 10.86 ± 17.14 | | 41.14 ± 13.52 | 49.89 ± 6.42 | 43.70 ± 6.38 | | 94.80 ± 1.93 | 96.00 ± 6.93 | 95.29 ± 3.27 | |
| | Stop | 53.84 ± 33.95 | 24.8 ± 26.44 | 28.22 ± 21.34 | | 51.00 ± 40.68 | 22.27 ± 34.53 | 16.13 ± 12.52 | | 96.10 ± 3.85 | 92.00 ± 6.93 | 93.81 ± 2.36 | |
| 4×4 | Ped-Xing | 20.11 ± 27.44 | 14.40 ± 17.34 | 14.37 ± 15.89 | 32.8 ± 26.18 | 10.00 ± 22.36 | 0.80 ± 1.79 | 1.48 ± 3.31 | 43.2 ± 32.78 | 95.99 ± 3.86 | 90.67 ± 4.62 | 93.13 ± 1.30 | 92.8 ± 2.02 |
| | School Zone | 32.74 ± 13.83 | 73.60 ± 29.07 | 39.85 ± 4.89 | | 53.72 ± 7.01 | 72.80 ± 12.13 | 61.22 ± 5.22 | | 89.07 ± 6.64 | 94.67 ± 2.31 | 91.69 ± 3.78 | |
| | Signal Ahead | 50.86 ± 9.26 | 43.20 ± 32.55 | 40.94 ± 20.42 | | 37.07 ± 4.29 | 75.20 ± 11.10 | 49.29 ± 3.90 | | 93.73 ± 5.69 | 94.67 ± 2.31 | 94.09 ± 1.76 | |
| | Speed Limit-45 | 24.11 ± 21.15 | 9.60 ± 10.81 | 13.54 ± 14.39 | | 39.37 ± 9.94 | 48.00 ± 16.25 | 41.82 ± 8.92 | | 91.70 ± 7.31 | 93.33 ± 8.33 | 92.07 ± 0.68 | |
| | Stop | 52.59 ± 35.62 | 23.20 ± 28.90 | 25.45 ± 24.02 | | 90.00 ± 14.14 | 19.20 ± 8.67 | 30.56 ± 11.57 | | 95.76 ± 0.28 | 90.67 ± 6.11 | 93.08 ± 3.39 | |
| 5×5 | Ped-Xing | 62.21 ± 41.64 | 10.40 ± 8.29 | 13.76 ± 7.14 | 35.04 ± 29.72 | 10.00 ± 22.36 | 0.80 ± 1.79 | 1.48 ± 3.31 | 42.88 ± 34.74 | 96.15 ± 6.66 | 92.00 ±4.00 | 93.92 ± 3.89 | 93.6 ± 2.56 |
| | School Zone | 33.44 ± 15.81 | 81.60 ± 26.17 | 42.24 ± 7.15 | | 51.93 ± 10.91 | 72.00 ± 5.66 | 59.76 ± 7.09 | | 91.09 ± 1.59 | 94.67 ±6.11 | 92.75 ± 2.56 | |
| | Signal Ahead | 70.38 ± 19.19 | 44.00 ± 31.62 | 44.92 ± 18.26 | | 38.20 ± 3.65 | 80.00 ± 7.48 | 51.52 ± 3.44 | | 96.05 ± 3.85 | 93.33 ±2.31 | 94.61 ± 1.07 | |
| | Speed Limit-45 | 42.81 ± 36.41 | 9.60 ± 15.13 | 12.19 ± 16.94 | | 39.58 ± 8.15 | 47.20 ± 14.25 | 41.98 ± 7.36 | | 89.31 ± 5.80 | 97.33 ±2.31 | 93.05 ± 2.75 | |
| | Stop | 58.98 ± 39.10 | 29.60 ± 28.37 | 32.39 ± 26.73 | | 90.95 ± 13.08 | 14.40 ± 6.69 | 24.04 ± 10.08 | | 97.16 ± 2.46 | 90.67 ± 2.31 | 93.79 ± 2.09 | |

Prec: Precision; Rec: Recall; Acc: Accuracy; Std: Standard Deviation

### 6.5 **Computation time**

To evaluate the feasibility of our defense strategy for real-time implementation, we measure the additional computation time required to reconstruct each adversarial patch-free image, as well as the computation time for traffic sign classification without and with the defense strategy. This assessment helps determine the computational overhead introduced by our defense strategy compared to a traffic sign classification system without it. Fig. 10 presents the patch-free image reconstruction time and the computation time for traffic sign classification, with and without a defense strategy. The average computation times for image reconstruction, classification without defense, and classification with defense are 0.9 ms, 4.13 ms, and 5.04 ms, respectively. The maximum values recorded for these processes are 1.51 ms, 5 ms, and 6.54 ms, respectively. The results demonstrate that the additional computational latency introduced by the defense strategy is negligible compared to that of a traffic sign classification system without any defense mechanism. This highlights the feasibility of our method for real-time implementation in autonomous driving.

We also compare the computational time of our method with the PAD method, and the results are summarized in Table 6. The computation time highlights a substantial computational advantage for our defense strategy over the PAD baseline. The average defense time for our method is approximately 0.91 ms, which is several orders of magnitude lower than that of PAD (1453.32 ms), indicating a significant reduction in computational overhead.

When considering end-to-end performance, the average traffic sign classification time with our defense is 5.04 ms, which lies well within the threshold latency for real-time safety applications in autonomous driving, 100 ms [21], compared to 1458.04 ms for PAD. In contrast, the baseline classification time without any defense remains comparable at approximately 4.61 ms. The narrow range between the minimum and maximum execution times for our method further underscores its robustness to input variation.

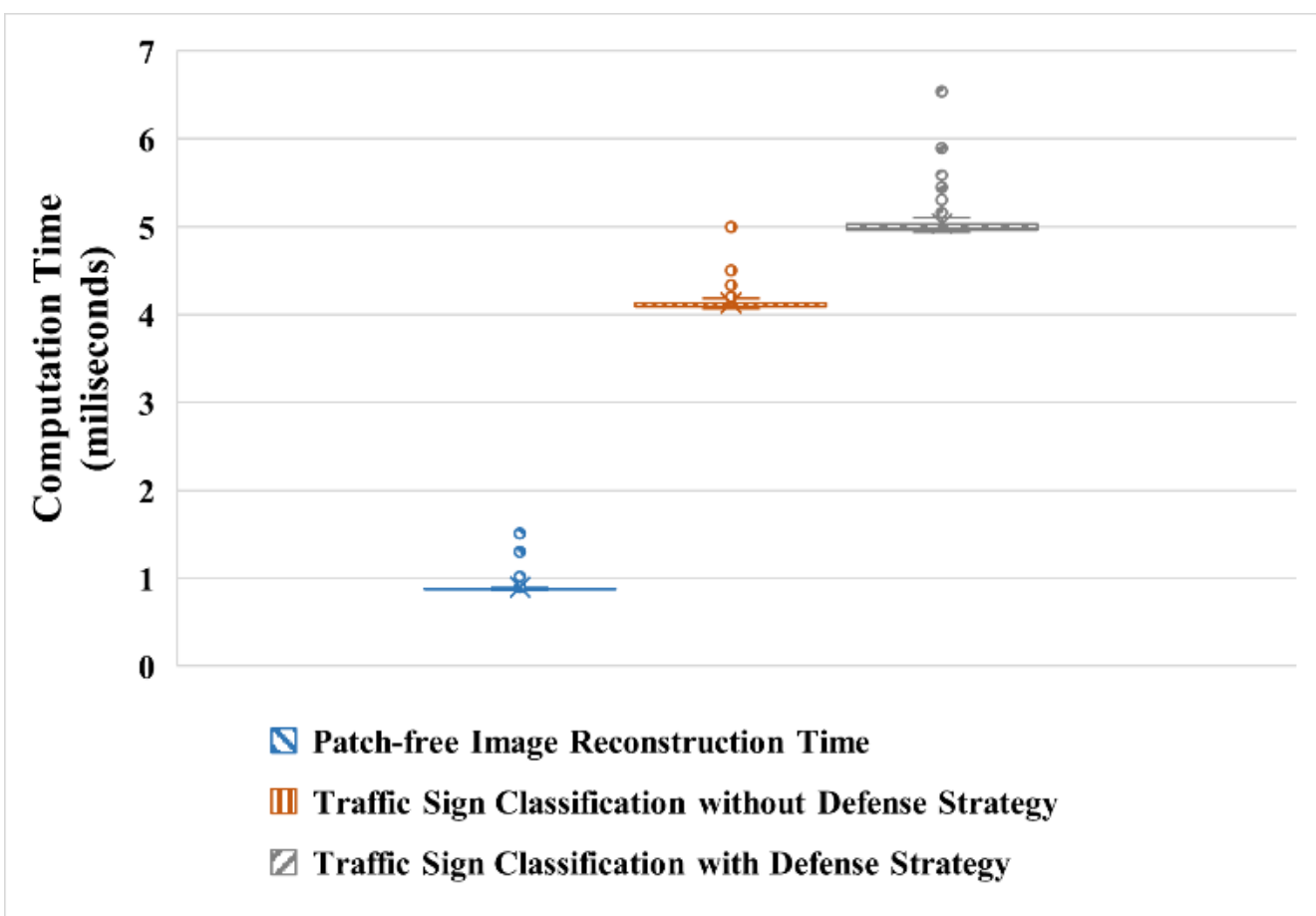

**Fig. 10** Computation time of traffic sign classification with and without defense strategy, along with patch-free image reconstruction time.

**Table 6** Comparison of computation time with benchmark method

| | Defense Time (PAD) (ms) | Defense Time (Ours) (ms) | Classification Time (ms) | Overall Latency (PAD) (ms) | Overall Latency (Ours) (ms) |
|---|---|---|---|---|---|
| **Min** | 1301.7 | 0.86 | 4.5 | 1306.3 | 4.94 |
| **Max** | 1856.2 | 1.51 | 5.1 | 1861 | 6.54 |
| **Avg** | 1453.32 | 0.9 | 4.61 | 1458.04 | 5.04 |
| **Std** | 107.72 | 0.08 | 0.13 | 107.76 | 0.19 |

Min: Minimum; Max: Maximum; Avg: Average; Std: Standard Deviation

The higher latency observed for the PAD method can be attributed to its reliance on two general properties of adversarial patches: spatial heterogeneity, where the patch exhibits texture or visual characteristics that differ from the surrounding regions of the image, and semantic independence, where the patch behaves as a context-independent element that does not conform to the semantics of the scene. Leveraging these cues, PAD performs a localization step, often based on recompression or mutual-information–driven analysis, to isolate the anomalous region, which is then removed or masked before the image is passed to the downstream object detector. This pipeline introduces substantial latency because it relies on iterative, computationally intensive scanning and reconstruction procedures to first detect the patch and then modify the image, rather than a single, fast forward pass through a trained model.

Overall, these results indicate that our strategy not only enhances robustness against APAs but also achieves real-time performance, making it well-suited for deployment in time-critical AV perception systems.

## 7. **Evaluation on Benchmark Dataset**

In this study, we conduct additional experiments on a benchmark dataset to evaluate the effectiveness of our defense strategy in scenarios involving diverse image classes with substantial sample counts per class. This evaluation ensures that the defense strategy not only generalizes beyond our customized dataset but is also robust and reproducible for real-world implementations, even as the number of different image classes increases. We chose the MNIST handwritten digit dataset for this evaluation, which contains 70,000 images. First, we train the ResNet-50 model on the MNIST dataset and then test it. The overall detection accuracy is 99.04%.

Second, we generate adversarial patches of sizes 3×3, 4×4, and 5×5 using the method described in Section 3. After performing APAs on the MNIST images, the average accuracy drops to 12.24% across different adversarial patch sizes, with class-wise accuracy under attack shown in Fig. 11.

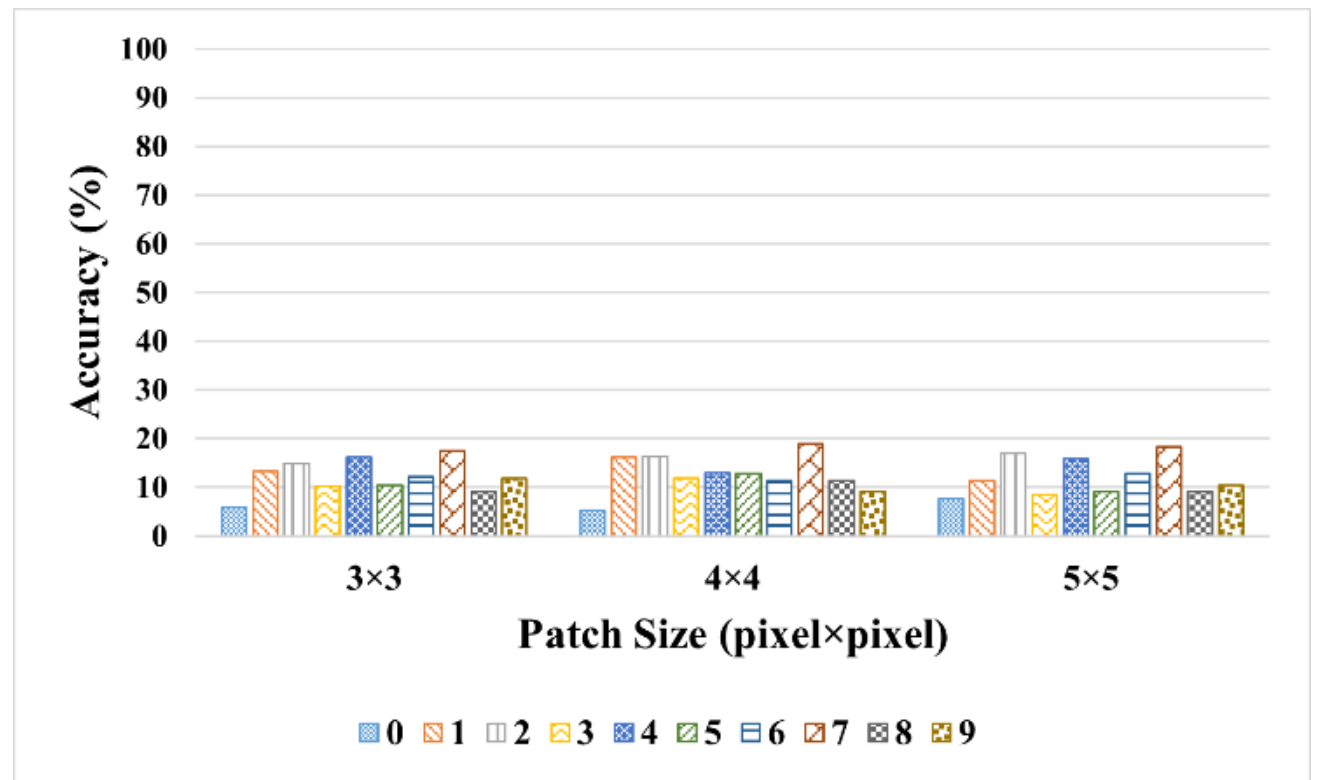

**Fig. 11** Accuracy of the ResNet-50 model for individual classes of the MNIST dataset under attack conditions for different patch sizes of APAs.

Third, we train the GAN following the approach in Section 4 to restore patch-free images. To ensure robustness, we randomly place patches (not the actual adversarial patches) of varying sizes and positions on MNIST training images, creating a patched training dataset, as described in Section 4. The GAN is trained using these sets of patched and respective unpatched images to reconstruct clean images.

Finally, we apply adversarial patches to the test images, pass them through the generator for reconstruction, and then classify them using the ResNet-50 model. Fig. 12 presents the accuracy improvements achieved by incorporating this defense strategy. The accuracy increases from 12.01% to 96.85%, from 12.61% to 94.11%, and from 12.00% to 87.58% under APAs for adversarial patches of size 3×3, 4×4, and 5×5, respectively, demonstrating that our defense strategy is generalizable, robust, and reproducible for real-world setups, even when applied to an increased number and different types of image classes.

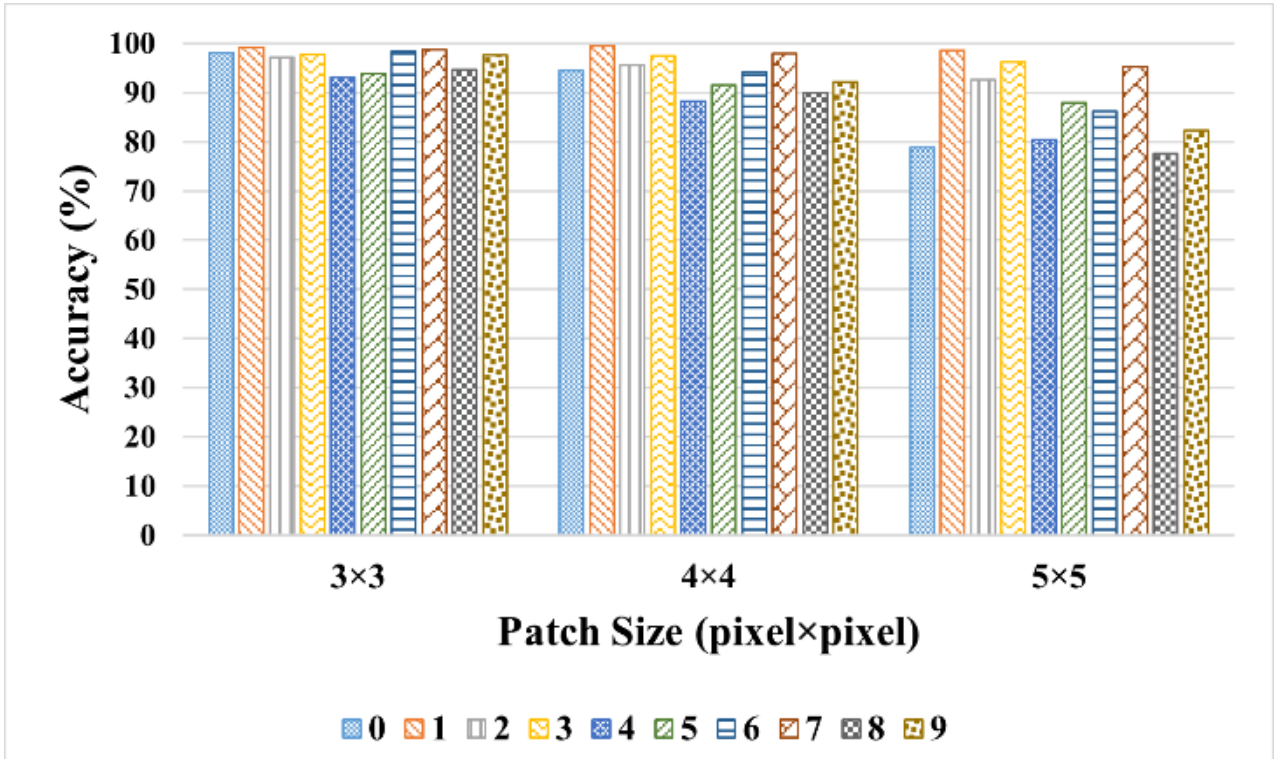

**Fig. 12** Accuracy of the ResNet-50 model for individual classes of the MNIST dataset under attack conditions for different patch sizes of APAs, after our defense strategy is incorporated.

## 8. Conclusions

In this study, we develop a GAN-based single-stage defense strategy to protect traffic sign classification against APAs. Our single-stage approach effectively neutralizes the impact of APAs in a single step, unlike existing methods that rely on patch detection followed by mitigation, which often introduce higher computational overhead.

Our strategy improves traffic sign detection accuracy against adversarial patches of varying sizes. A single generator can reconstruct patch-free traffic sign images across multiple classes, making our approach both efficient and scalable. Furthermore, our strategy can remove adversarial patches from traffic sign images without prior knowledge of the specific patches used.

Additionally, our strategy is adaptable to different traffic sign classifiers that use various deep learning (DL) models. During GAN training, classification loss and perceptual loss are computed based on the target classifier's DL model, making the defense strategy applicable to other classification models. This approach paves the way for APA-resilient, real-time traffic sign classification systems for autonomous vehicles and other related intelligent transportation applications.

Finally, we utilize the MNIST benchmark dataset to further evaluate our GAN-based defense strategy. Our analysis indicates that our defense strategy increases classification accuracy on the MNIST dataset under APA conditions, demonstrating that our method is robust and reproducible for real-world applications.

## Acknowledgment

This work is based upon the work supported by the National Center for Transportation Cybersecurity and Resiliency (TraCR) (a US Department of Transportation National University Transportation Center) headquartered at Clemson University, Clemson, South Carolina, USA. Any opinions, findings, conclusions, and recommendations expressed in this material are those of the author(s) and do not necessarily reflect the views of TraCR, and the US Government assumes no liability for the contents or use thereof.

## Author Contributions

Conceptualization, A.E. and M.C.; methodology, A.E.; software, A.E.; validation, A.E. and M.C.; formal analysis, A.E.; investigation, A.E. and M.C.; data curation, A.E.; writing-original draft preparation, A.E. and M.C.; writing-review and editing, A.E. and M.C.; visualization, A.E.; supervision, M.C. All authors have read and agreed to the published version of the manuscript.

## Funding

This research was funded by the National Center for Transportation Cybersecurity and Resiliency (TraCR) (a US Department of Transportation National University Transportation Center) headquartered at Clemson University, Clemson, South Carolina, USA, under Grants: 69A3552344812, 69A3552348317.

## Declarations

**Conflict of interest** The authors declare that they have no conflict of interest.